\documentclass[11pt,a4paper]{article}
\usepackage[hyperref]{acl2021}
\usepackage{times}
\usepackage{latexsym}

\usepackage{microtype}

\aclfinalcopy 

\newcommand{\protoaugment}{\textsc{ProtAugment}}
\usepackage{amssymb}
\usepackage{amsmath}
\usepackage{graphicx}

\usepackage{booktabs}
\newcommand{\ra}[1]{\renewcommand{\arraystretch}{#1}}
\usepackage{multirow, makecell}
\usepackage{rotating}
\usepackage{array, caption}
\usepackage{tabularx}

\usepackage{soul}
\usepackage{color}

\usepackage{dblfloatfix, placeins}
\usepackage{natbib}

\newcommand\Mark[1]{\textsuperscript#1}

\title{\protoaugment{}: Unsupervised diverse short-texts paraphrasing for intent detection meta-learning}

\author{
    Thomas Dopierre\Mark{1}\Mark{2} \and Christophe Gravier\Mark{1} \and \\\and \textbf{Wilfried Logerais\Mark{2}}
\\ \\
\begin{tabular}{*{2}{>{\centering}p{.5\textwidth}}}
\Mark{1}Laboratoire Hubert Curien\\UMR CNRS 5516\\Université Jean Monnet&   {\Mark{2}Meetic \\Paris, France} \tabularnewline
Saint-Étienne, France &  \tabularnewline
\url{firstname.lastname@univ-st-etienne.fr} & \url{{t.dopierre,w.logerais}@meetic-corp.com}
\end{tabular}
}

\date{}

\begin{document}
    \maketitle
    \begin{abstract}
        Recent research considers few-shot intent detection as a meta-learning problem: the model is \textit{learning to learn} from a consecutive set of small tasks named episodes.
%
%
        In this work, we propose \protoaugment{}, a meta-learning algorithm for short texts classification applied to the intent detection task.
        \protoaugment{} is a novel extension of Prototypical Networks~\cite{snell2017prototypical} that limits over-fitting on the bias introduced by the few-shots classification objective at each episode.
        It relies on diverse paraphrasing: a conditional language model is first fine-tuned for paraphrasing, and diversity is later introduced at the decoding stage at each meta-learning episode.
        The diverse paraphrasing is unsupervised as it is applied to unlabelled data and then fueled to the Prototypical Network training objective as a consistency loss.
        \protoaugment{} is the state-of-the-art method for intent detection meta-learning, at no extra labeling efforts and without the need to fine-tune a conditional language model on a given application domain.
        %

%
%
%
%
    \end{abstract}

    \section{Introduction}\label{sec:intro}

    Intent detection, a sub-field of text classification, involves classifying user-generated short-texts into intent classes, usually for conversational agents applications~\cite{Casanueva2020}.
    Since conversational agent applications are domain-specific, intent detection is a challenging task because of labeled data scarcity and the number of classes (intents) it usually involves~\cite{Dopierre20}.
    As a consequence, recent research~\cite{snell2017prototypical, protopp} considers few-shot intent detection as a
    meta-learning problem: the model is trained to classify user utterances from a consecutive set of small tasks
    named episodes.
    Each episode contains a limited number of $C$ classes alongside a limited number of $K$ labeled data for each of
    the $C$ classes -- this is usually referred to as a $C$-way $K$-shots setup.
    At test time, the algorithm is evaluated on classes that were not seen during training.
%
%
    That is the reason why meta-learning is sometimes referred to as \textit{learning to learn}: it mimics human abilities to learn iteratively from different and small tasks.
%
%
    Meta-learning has successfully been applied to a wide set of NLP tasks: hypernym detection~\cite{Yu20}, low resource machine translation~\cite{Gu18}, machine understanding tasks~\cite{Dou19} or
    structured query generation~\cite{Huang18}.
%
%
%
    Most meta-learning algorithms (Section~\ref{sec:related}) were developed in the course of the last 5 years. It has recently been empirically demonstrated that comparative studies in follow-up papers of~\cite{snell2017prototypical} are debatable -- for short texts classification -- because of the two following main issues~\cite{Dopierre21}.
    %
%
%
First, comparative studies involve simple and limited datasets in terms of number and separability of classes (SNIPS~\cite{snips}, a very popular dataset, includes only $7$ classes, with the current best model performing over 99\% accuracy~\cite{cao2020balanced}).
%
%
Second, as we further better understand~\cite{Niven19}, fine-tune~\cite{Liu19R,Hao20} and refine~\cite{Khetan20}
BERT-derived models, it is not clear if the different meta-learning frameworks can be considered state-of-the-art
due to their architecture or due to the improvements of available text encoders at the time of conception.
\cite{Dopierre21} concludes that Prototypical Networks~\cite{snell2017prototypical} (that were using LSTM-based text encoders when introduced in NLP) are actually the state-of-the-art for intent detection when equipped with a fine-tuned BERT text encoder model.
%
%
Ultimately, improving Prototypical Networks have therefore been proven to be a very challenging task in reality. \\
\indent A cornerstone challenge is that meta-learning models can easily overfit on the biased distribution introduced by a few training examples~\cite{Yang2021}.
In order to prevent overfitting and inspired by~\cite{Xie20}, we introduce an unsupervised diverse paraphrasing loss in the Prototypical Networks framework.
A key idea is consistency learning: by augmenting unlabeled user utterances, \protoaugment{} enforce a more robust text representation learning.
Unfortunately, back-translation is a poor data augmentation strategy for short-texts: neural machine translation provides very similar (if not the same) sentences to the original ones, which hinders its ability to provide diverse augmentations (Section~\ref{subsec:diversity}).
%
%
Consequently, in this work, we transfer a denoising autoencoder pre-trained on the sequence-to-sequence task~\cite{Lewis20} to the paraphrase generation task and then use it to generate paraphrases.
%
%
%
%
As fine-tuning is very efficient for such a model, it is not easy to optimize it for diverse paraphrasing.
\cite{Goyal20} presents an approach for diverse paraphrasing that reorders the original sentence to guide the conditional language model to generate diverse sentences. The diversity in that work is provided by the reordering of the elements, which surprisingly affects the attention mechanism.
In~\cite{Liu20}, expression diversity is part of the unsupervised paraphrasing system supported by simulated annealing.
Both approaches imply domain transfer, and consequently, as many diverse paraphrasing models to maintain as the number of considered application domains, which do not scale very well.
%
%
In this work, we instead introduce diversity in the downstream decoding algorithm used for paraphrase generation.
%
%
Diverse decoding methods are mostly extensions to the beam search algorithm, including noise-based algorithms~\cite{cho2016noisy}, iterative beam search~\cite{Kulikov19}, clustered beam search~\cite{Tam20} and diverse beam search~\cite{Ashwin18}.
%
%
There is no clear optimal solution, the choice is task-specific and dependent on one's tolerance for lower quality outputs as a diversity/fluency trade-off~\cite{Ippolito19}.
%
%
While diverse beam search allows controlling the diversity/fluency trade-off partially, we further demonstrate that adding constraints to diverse beam search in order to generate tokens not seen in the input sentence (that is, \textit{constrained diverse beam search}) is a simple yet powerful strategy to further improve the diversity of the paraphrases.
%
%
%
%
Paired with paraphrasing user utterances and its consistency loss incorporated in Prototypical networks, our model is the best method for intent detection meta-learning on 4 public datasets, with neither extra labeling efforts nor domain-specific conditional language model fine-tuning.
We also show that \protoaugment{}, having access to only $10$ samples of each class of the training data, still significantly outperforms a Prototypical Network which is given access to \textit{all} samples of the same training data.
%
%
%
%
%

\section{Neural architectures for meta-learning}\label{sec:related}
%
%
%
Past works on meta-learning for classification tasks investigate how to best predict a query point's class at an episode scale.
%
This process is bounded to the set of the $C$ classes considered in a given episode.
Matching Networks~\cite{vinyals2016matching} predict the class of a query point as the average cosine distance between the query vector and all support vectors for each class.
%
%
%
%
Prototypical Networks~\cite{snell2017prototypical} extend Matching Networks: after obtaining support vectors from the encoder, a class \textit{prototype} is produced via a class-wise vector averaging operation.
All query points are then predicted with respect to their distance (cosine or euclidean) to all prototypes.
Like Prototypical Networks, Relation Networks~\cite{Sung_2018_CVPR} emerged from Computer Vision application and were later successfully applied to NLP~\cite{Zhang2018}.
They introduce a relation module, which captures the relationship between data points: instead of using a pre-defined distance (euclidean or cosine most of the time), this approach allows such networks to learn this metric by themselves.
This is achieved using either a shallow feed-forward sub-network or a Neural Tensor Layer relation module~\cite{Socher13} (intermediate learnable matrices).
Another extension to Prototypical Networks is provided in~\cite{protopp}.
Unlabeled data are incorporated using two distinct approaches: i) taking unlabeled data from the same classes as the episode or ii) using any unlabeled data and incorporating both a distractor cluster and masking strategy to minimize the impact of distant unlabeled points.
The first approach is unrealistic for meta-learning, as it implies knowing the unlabeled data class.
The second method assumes that all the noise is centered around a single distractor cluster and introduces an additional hyperparameter for masking -- which is hardly fine-tuneable for small few-shot datasets.

\section{Background}\label{sec:background}

\subsection{Notations}\label{subsec:notations}
Meta-learning algorithms are trained using a specific procedure made of consecutive episodes.
Let $\mathcal{C}_{ep}$ be the set of $C$ classes sampled for the current training episode, such as
$\mathcal{C}_{ep} \subset \mathcal{C}_{train}$, where $\mathcal{C}_{train}$ is the set of all classes available
for training.
We note $\mathcal{C}_{test}$, the set of classes used for testing, with $\mathcal{C}_{train} \cap
\mathcal{C}_{test} = \emptyset$.
Each class $c \in \mathcal{C}_{ep}$ comes with $K$ labeled samples, used as support.
The set of $C \times K$ samples are usually referred to as $\mathcal{S}$, the support set, so that $\mathcal{S} = \{
(x_1, y_1),\dots,(x_{C \times K}, y_{C \times K})\}$.
We denote $S_{c}$ the set of support examples labeled with class $c$.
Each episode comes with a query set $\mathcal{Q}$, which serves as the episode-scale optimization -- the
model parameters are updated based on the prediction loss on $\mathcal{Q}$, given $\mathcal{S}$ as an input.
$\mathcal{Q}_c$ is the set of query examples labeled with class $c$.
%
%
%

    \subsection{Prototypical networks}\label{subsec:protobg}

    In prototypical networks, each class is mapped to a representative point, called \textit{prototype}.
    %
    %
    Each sample is first encoded into a vector using an embedding function $f_{\phi}$ with learnable parameters $\phi$ -- this is the function we want to optimize.
    Using these embeddings, we compute each prototype $p_c, c \in \mathcal{C}_{ep}$ as the mean vector of embedded support points belonging to the class $c$, as described in Equation~\ref{eq:prototypes}.
    \begin{equation}
        \label{eq:prototypes}
        p_{c} = \frac{1}{K}\sum_{(x_i, y_i) \in S_{c}}f_{\phi}(x_i)
    \end{equation}
    Given those prototypes and a distance function $d$, prototypical networks assign a label to a query point by computing the softmax over distances between this point's embedding and the prototypes, as in Equation~\ref{eq:prototypical-network-1}. In the original paper, \cite{snell2017prototypical} use the euclidean distance and we also observed consistent slightly worse results with the cosine distance. 
%
    \begin{equation}
        \label{eq:prototypical-network-1}
        \resizebox{.8 \linewidth}{!}{$\mathbb{P}_{\phi}(y=c | x) ={\text{softmax}\left(-d(f_{\phi}(x), p_c)\right)}$}
    \end{equation}
%
    The supervised loss function $\bar{L}$ is the average negative log-probability of the correct class assignments for all query points.
    At test time, episodes are created using classes from $\mathcal{C}_{test}$, and accuracy is measured as the query points assignments, given prototypes derived from the support points.

%



    \section{\protoaugment{}}\label{sec:pa}
    In this section, we present our semi-supervised approach \protoaugment. Along with the labeled data randomly chosen at each episode, this approach uses $U$ unlabeled data randomly drawn from the whole dataset -- that is, data from training, validation, and test labels. We first do a data augmentation step from this unlabeled data, where we obtain $M$ paraphrases for each unlabeled sentence. The $m^{th}$ paraphrase of $x$ will be denoted $\tilde{x}^m$. Then, given unlabeled data and their paraphrases, we compute a fully unsupervised loss. Finally, we combine both the supervised loss $\bar{L}$ (the Prototypical Network loss using labeled data) and unsupervised loss (denoted $\tilde{L}$) and run back-propagation to update the model's parameters.

\begin{figure*}[h!]
    \centering
    \includegraphics[width=\textwidth]{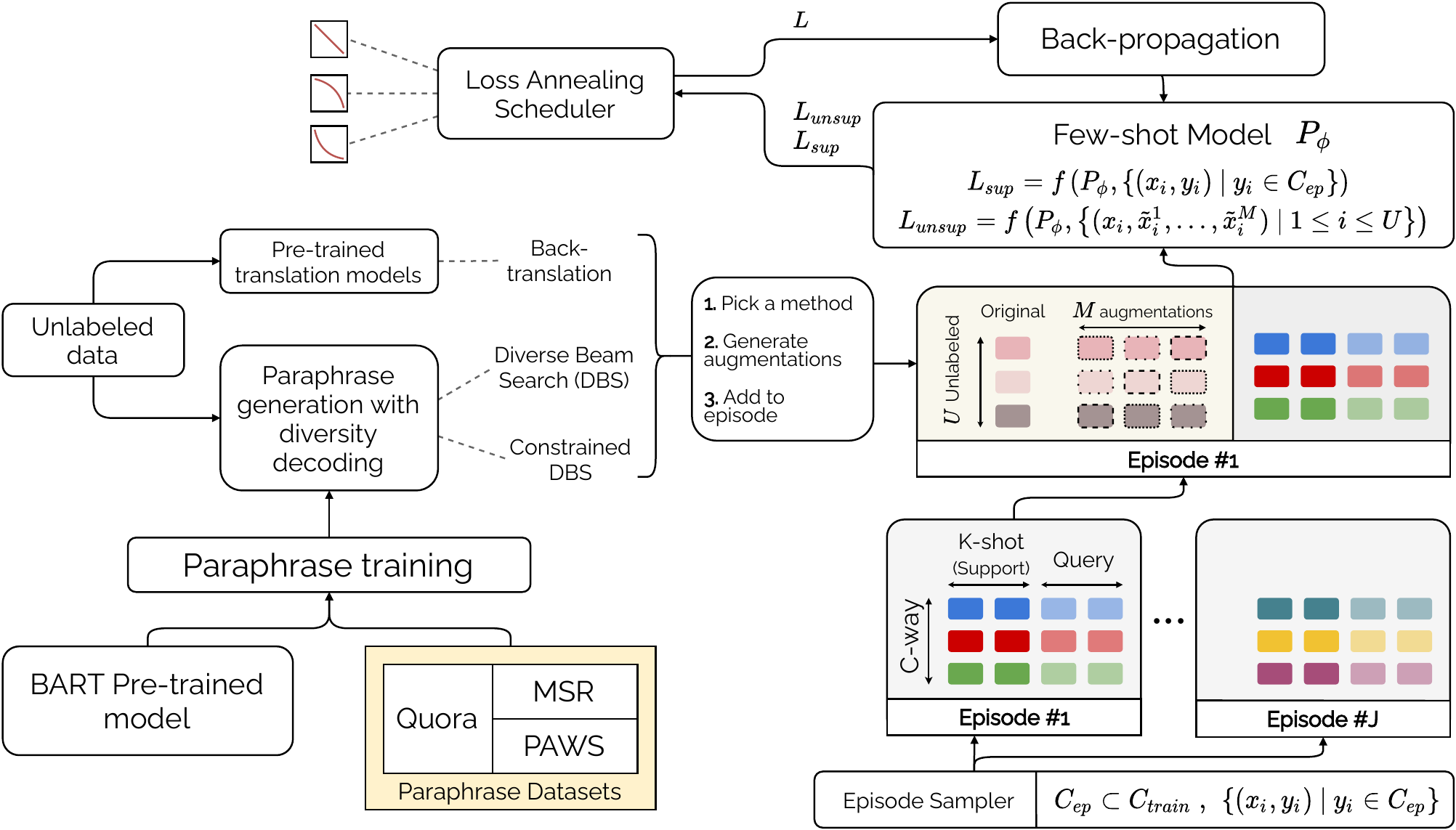}
    \caption{\protoaugment{} illustrated on a 3-way 2-shot short text classification meta-learning task ($C=3$, $K=2$). BART is pre-trained for the paraphrasing task on three datasets: Quora~\cite{Sharma19}, MSR~\cite{Zhao10} and Google PAWS-Wiki~\cite{Yang19,Zhang19}. The paraphrase model is used to paraphrase unlabeled samples but equipped with diversity strategies (back translation being proposed as a baseline). The final loss is computed using a loss annealing scheduler, which is expected to smooth the supervised (given shots) and unsupervised (augmented unlabeled sentences) prediction errors to yield parameter gradients. A new episode means sampling other classes along with their support and query points.
}
    \label{fig:protoaugment}
\end{figure*}


    \subsection{Generating augmentations through paraphrasing}\label{subsec:pa:paraphrasing}

    The BART~\cite{Lewis20} model is a Transformer-based neural machine translation architecture that is trained to remove artificially corrupted text from the input thanks to an autoencoder architecture.
    While it is trained to reconstruct the original noised input, it can be fine-tuned for task-specific conditional generation by minimizing the cross-entropy loss on new training input-output pairs~\cite{Bevilacqua20}.
    In \protoaugment{}, we fine-tune a pre-trained BART model on the paraphrasing task. %
    %
    The paraphrase sentence pairs we use for this task are taken from 3 different paraphrase detection
    datasets\footnote{we take only pairs that are paraphrases of each other since these are \textit{paraphrase
    detection} datasets}: Quora~\cite{Sharma19}, MSR~\cite{Zhao10}, and Google PAWS-Wiki~\cite{Yang19,Zhang19}. Those
    datasets have different sizes, and the largest one -- Quora -- consist of 149,263 pairs of duplicate questions. To
    balance turns of sentences (questions/non questions paraphrases), $50\%$ of our fine-tuning paraphrase datasets is made of Quora, $5.6\%$ of MSR and $44.4\%$ PAWS-Wiki. This yields 94,702 sentence pairs to train the model on the paraphrasing task. We include both code and data on our github repository~\footnote{https://github.com/tdopierre/ProtAugment}.
    %

    Using this fine-trained paraphrasing model, we can generate paraphrases of unlabeled sentences, hopefully having paraphrases representing the same intents as the original sentences.
    To add some diversity in the generated paraphrases, we use Diverse Beam Search (DBS) instead of the regular Beam Search. As~\citet{Ashwin18} has shown in the original paper, adding a dissimilarity term during the decoding step helps the model produce sequences that are quite far from each other while still retaining the same meaning. The next section describes how we constrained this decoding to enforce even more diversity among generated paraphrases in \protoaugment{}.

    \subsection{Constrained user utterances generation}\label{subsec:pa:constraienddbs}
    While DBS enforces diversity between the generated sentences, it does not ensure diversity between the generated paraphrases and the original sentences. It was formerly designed for tasks that do not need this diversity with the original sentence  (translation, image captioning, question generation). To enforce that our generated paraphrases are diverse enough, we further constraint DBS by forbidding using parts of the original sentences. In the following paragraphs, we introduce two forbidding strategies.

    \paragraph{Unigram Masking.} In this strategy, we randomly select tokens from the input sentence which will be forbidden at the generation step. The goal here is to force the model to use different words in the generated sentences than it saw in the original sentences. Each word of the input sentence is randomly masked using a probability $p_{\text{mask}}$.
    The underlying assumption is that forbidding tokens at the beginning of a sentence with a higher probability than the end of the sentence may have a greater impact on the beam search algorithm. Indeed, as the decoding is a conditional task based on prior generated tokens, masking the first tokens may significantly impact diversity.
    %
    %
    \begin{table*}[!b]
        \ra{0.8}
        \centering
        \resizebox{0.85\linewidth}{!} {%
            \begin{tabular}{
                @{\hskip 0.18in}
                c@{\hskip 0.28in}
                c@{\hskip 0.28in}
                c@{\hskip 0.28in}
                c@{\hskip 0.28in}
                c@{\hskip 0.18in}
                }
                \bottomrule
                Dataset & \thead{\#sentences} & \thead{\#classes \\ train/valid/test (total)} & \thead{Available \\ sentences/class} &
                \thead{\#tokens/sentence}
                \\
                \toprule
                Banking77 & $13,083$ & $25/25/27(77)$ & $170\pm33$ & $11.7\pm7.6$ \\
                HWU64 & $11,036$ & $23/16.4/24.6(64)$ & $172\pm40$ & $6.6\pm2.9$ \\
                Clinic150 & $22,500$ & $50/50/50 (150)$ & $150\pm0$ & $8.5\pm3.3$ \\
                Liu & $25,478$ & $18/18/18 (54)$ & $472\pm831$ & $7.5\pm3.4$ \\

                \bottomrule
            \end{tabular}
        }

        \caption{
            Main statistics of intent detection evaluation datasets. For HWU64, each split's number of classes varies at each run to ensure there is no cross-split domain, hence the decimal number.
        %
        }
        \label{tab:datasets-stats}
    \end{table*}
    We therefore introduce two additional variants: one where we put more probability on the first tokens and the reverse where there is more weight in the last tokens. To ensure that all three variants mask the same amount of tokens on average, we ensure the area under the curve of the three probability functions are equal to a fixed value noted $p_{\text{mask}}$.
    %
%

    \paragraph{Bi-gram Masking} Another strategy we consider is to prevent the paraphrasing model from generating the same bi-grams as in the original sentence. This time, we are not masking any single word but forcing the model to change the sentence's structure, which will, hopefully, increase the diversity of the generated paraphrases.

    \subsection{Unsupervised diverse paraphrasing loss}\label{subsec:pa:unsupervised-loss}
    After generating paraphrases for each unlabeled sentence, we create unlabeled prototypes. For each unlabeled sentence $x_u \in U$, we derive the unlabeled prototype $p_{x_u}$ as the average embedding of the paraphrases of $x_u$ (Equation~\ref{eq:unlabeled-prototypes}).

    \begin{equation}
    \label{eq:unlabeled-prototypes}
        p_{x_{u}} = \frac{1}{M}\sum_{m=1}^{M} f_{\phi}(\tilde{x}^{m}_{u})
    \end{equation}

    After obtaining the unlabeled prototypes, we compute the distances between all unlabeled samples and all unlabeled prototypes. Given such distances, we model the probability of each unlabeled sample being assigned to each unlabeled prototype (Equation~\ref{eq:unlabeled-prototypes-proba}), as in the supervised part of the Prototypical Networks -- except this time, it is fully unsupervised. This probability should be close to 1 between an unlabeled sample and its associated unlabeled prototype and close to 0 otherwise.

    \begin{equation}
        \label{eq:unlabeled-prototypes-proba}
        \resizebox{.87\linewidth}{!}{$\mathbb{P}_{\phi}(u = v | x_u ) = {\text{softmax}\left(-d(f_{\phi}(x_u), p_{x_{v}})\right)}$}
    \end{equation}

    Given assign probabilities between unlabeled samples and unlabeled prototypes, we can compute a fully unsupervised cross-entropy loss $\tilde{L}$, training the model to bring each sentence closer to its augmentations' prototype and further from the prototypes of other unlabeled sentences.
    Recall that $f_{\phi}$ is the embedding function with $\phi$ as learnable parameters (Section~\ref{subsec:protobg}).

    After obtaining both supervised loss $\bar{L}$ and unsupervised loss $\tilde{L}$, we combine them into the final loss $L$ using a loss annealing scheduler (see Equation~\ref{eq:loss}), which will gradually incorporate the unsupervised loss as training progresses.
    \begin{equation}
        \label{eq:loss}
        L = t^{\alpha} \times \tilde{L} + \left(1 - t^{\alpha}\right) \times \bar{L} \quad ; \quad t \in \left(0, 1\right)
    \end{equation}

    The goal here is to mainly use the supervised loss first so that the model gets a sense of the classification task. Then, incorporating more and more knowledge from unlabeled samples will make the model more robust to noise, which is essential as it is constantly tested on classes it has never seen before. We explore three different strategies for gradually increasing the unsupervised contribution: a linear approach ($\alpha = 1$), an aggressive one ($\alpha = 0.25$), and a conservative one ($\alpha = 4$).

\FloatBarrier

    \section{Experiments}\label{sec:xps}

    \subsection{Datasets}\label{subsec:datasets}
    %
    We consider the DialoGLUE benchmark~\cite{Mehri20}, a set of natural language understanding benchmark for task-oriented dialogue, which contains three datasets for intent detection: \texttt{Banking77}, \texttt{HWU64} and \texttt{Clinic150} -- the three datasets were already available prior the release of DialoGLUE.
    Additionally, we also consider the \texttt{Liu57} intent detection dataset, as it contains the same order of magnitude of intent classes and is user-generated as well.
    All datasets are public and in English.
\paragraph{Banking77} The \texttt{Banking77} dataset \cite{Casanueva2020} classifies $13,083$ user utterances related to into $77$ different intents. This dataset i) is specific to a single domain (banking) and ii) requires a fine-grained understanding to classify due to intents being very similar. Following~\cite{Mehri20} and contrary to~\cite{Casanueva2020}, we designate a validation set along a training and a testing set for that dataset (Table~\ref{tab:datasets-stats}).
\paragraph{HWU64} \texttt{HWU64}~\cite{Liu19HWU64} classifies $25,716$ user utterances with $64$ user intents. It features intents spanning across $21$ domains (alarm, audio, audiobook, calendar, cooking, datetime, \ldots). When separating training, validation, and test labels, we ensure each domain is represented only in one set of labels. This ensures the model learns to discriminate between both intents and domains.
%
\paragraph{Clinic150} This dataset~\cite{OOS} classifies $150$ user intents in perfectly equally-distributed classes. This chatbot-like style dataset was initially designed to detect out-of-scope queries, though, in our experiments, we discard the out-of-scope class and only keep the $150$ labeled classes to work with, as in~\cite{Mehri20}.
\paragraph{Liu57} Introduced by~\citet{Liu-dataset}, this intent detection dataset is composed of $54$ classes. It was collected on Amazon Mechanical Turk, where workers were asked to formulate queries for a given intent with their own words. It is highly imbalanced: the most (resp. least) common class 
holds $5,920$ (resp. $24$) samples 

    \subsection{Experimental settings}\label{subsec:settings}
\textbf{Conditional language model and language model}. For the BART fine-tuning process, we used the defaults hyper-parameters reported in~\cite{Lewis20}, and we fine-tuned the BART model for a single epoch (two hours on a Titan RTX GPU). Increasing the number of epochs for fine-tuning BART degrades performances on the intent detection task: the downstream diverse beam search struggles to find diverse enough beam groups since the model perplexity has been lower with further fine-tuning (this is also hinted  in~\cite{Bevilacqua20}).
Our text encoder $f_{\phi}$ is a \texttt{bert-base} model, and the embedding of a given sentence is the last layer hidden state of the first token of this sentence. For each dataset, this model is fine-tuned on the masked language modeling task for 20 epochs. Then, the encoder of our meta learner is initialized using the weights of this fine-tuned model.

\paragraph{Datasets} From a dataset point-of-view, we create two data profiles: \textbf{full} (all the training dataset is available, the usual meta-learning scenario) and \textbf{low} (only 10 samples are available for each training class, an even more challenging meta-learning scenario in which a model meta-learns on very few samples per training class). All experimental setups are run 5 times. For each run, we randomly select training, validation, and testing classes, as well as the samples for the \textbf{low} setting. We train the few-shot models for a maximum of $10,000$ C-way K-shots episodes, evaluating and testing every $100$ episodes, stopping early if the evaluation accuracy has not progressed for at least $20$ evaluations. We evaluate and test using $600$ episodes, as in other few-shot works~\cite{snell2017prototypical,chen2019closerfewshot}.  We compare the systems in the following standard few-shot evaluation scenarios: $5$-way 1-shot, and $5$-way 5-shots.

\paragraph{Paraphrasing.} At each episode, we draw $U=5$ unlabeled samples to generate paraphrases from. For the back-translation baseline, we use the publicly available\footnote{https://huggingface.co/models?search=helsinki-nlp} translation models from the \texttt{Helsinki-NLP} team. We use the following pivot languages: \texttt{fr}, \texttt{es}, \texttt{it}, \texttt{de}, \texttt{nl}, which yields $5$ augmentations for each unlabeled sentence. For our experiments with Diverse Beam Search, we generate sentences using 15 beams, group them into $5$ groups of $3$ beams. In each group, we select the generated sentence which is the most different from the input sentence using BLEU as a metric for diversity. This yields $M=5$ paraphrases for each unlabeled sentence, as in the back-translation baseline. DBS uses a diversity penalty parameter to penalize words that have already been generated by other beams to enforce diversity. As advised in the original DBS paper~\cite{Ashwin18}, we set the diversity penalty to $0.5$ in our experiments, which provides diversity while limiting model hallucinations. Our Unigram Masking strategy's masking probability is set to $p_{\text{mask}}=0.7$ found by linear search from 0 to 1 with steps of $0.1$.

\begin{table}[ht]
 \ra{0.9}
\resizebox{1\linewidth}{!} {%
\begin{tabular}{l}
\toprule

\textit{orig}: How long will my transfer be pending for?\\
\textit{back}: How long will my transfer be on hold?\\
\textit{dbs\_0}: How long will my transfer be pending? I am in first year.\\
\textit{dsb\_1}: When are all transfers coming up and how many days are they expected?\\
\textit{dbs\_2}: If I have a transfer for a while, how long should I wait for it?\\
\midrule

\textit{orig}: I am not sure where my phone is.\\
\textit{back}: I don't know where my phone is.\\
\textit{dbs\_0}: I am not really sure where my phone is located\\
\textit{dsb\_1}: How can I find the location of any Android mobile\\
\textit{dbs\_2}: I don't know where is my cell phone\\
\midrule

\textit{orig}: can you play m3 file\\
\textit{back}: can you read m3 file\\
\textit{dbs\_0}: M3 files: can I play the entire M3 file?\\
\textit{dsb\_1}: Is there any way to play 3M files on Earth without downloading it\\
\textit{dbs\_2}: Is there any way to play M3 files on Windows?\\

\bottomrule
\end{tabular}
}
\caption{Examples of sentences (\textit{orig}) paraphrased using back translation (\textit{back}), vanilla diverse beam search -- DBS (\textit{dbs\_0}), DBS with unigram masking (\textit{dbs\_1}) and DBS with bigram masking (\textit{dbs\_2}).. }
\label{tab:samples}
\end{table}

\subsection{Evaluation of paraphrase diversity}\label{subsec:diversity}
We evaluate the diversity of paraphrases for each method, and report results for two representative datasets in
Table~\ref{tab:paraphrase-evaluation} (due to space limitations, the report for all datasets is given in
    appendix~\ref{app:paraphrase-diversity-evaluation}).
    For each paraphrasing method and each dataset, metrics are computed over unlabeled sentences and their paraphrases. To assess the diversity of paraphrases generated by the different methods, the popular BLEU metric in Neural Machine Translation is a poor choice~\cite{Bawden20}. We use the bi-gram diversity (\textbf{dist-2}) metric as proposed by~\cite{Ippolito19}, which computes the number of distinct 2-grams divided by the total amount of tokens. We also report the average similarity (denoted \textbf{use}) within each sentence set, using the Universal Sentence Encoder as an independent sentence encoder. Results show that paraphrases obtained with back-translation are too close to each other, resulting in a high sentence similarity and low bi-gram diversity. On the other hand, DBS generates more diverse sentences with a lower similarity. Our masking strategies strengthen this effect and yield even more diversity. The measured diversity strongly correlates with the average accuracy of the intent detection task~(Table~\ref{tab:results}).

\begin{table}[h]
    \centering
    \resizebox{\linewidth}{!} {%
    \begin{tabular}{@{\extracolsep{1pt}}lcccc}
    \toprule
         &  \multicolumn{2}{c}{BANKING77} & \multicolumn{2}{c}{HWU64} \\
         \cmidrule(lr){2-3}
         \cmidrule(lr){4-5}
         & \textbf{dist-2} & \textbf{use} & \textbf{dist-2} & \textbf{use} \\
         \midrule
         back-translation & 0.183 & 0.896 & 0.307 & 0.888\\
         DBS &  0.200 & 0.807 & 0.340 & 0.769 \\
         DBS+bigram & 0.228 & 0.702 & 0.350 & 0.692 \\
         DBS+unigram & \textbf{0.343} & \textbf{0.613} & \textbf{0.407} & \textbf{0.628} \\
    \bottomrule
    \end{tabular}
    }
    \caption{Paraphrase diversity measures. For \textbf{dist-2} (resp. \textbf{use}) higher values (resp. lower) indicates more diversity.}
    \label{tab:paraphrase-evaluation}
\end{table}

\begin{table*}[h!]
 \ra{1.1}
\centering

\resizebox{\linewidth}{!} {%
\begin{tabular}{@{\extracolsep{1pt}}llcccccccccc}
\toprule
{} & {} & \multicolumn{8}{c}{\textbf{Datasets}} & \multicolumn{2}{c}{\textbf{Accuracy stats}}\\
 \cmidrule(lr){3-10}

 {\multirow{2}{1.1cm}{\textbf{Data Profile}}} & {\multirow{2}{1.1cm}{\textbf{Method}}} & \multicolumn{2}{c}{Banking} & \multicolumn{2}{c}{HWU} & \multicolumn{2}{c}{Liu} & \multicolumn{2}{c}{Clinic} & \multicolumn{2}{c}{(\textbf{$AVG\pm STD$})}\\
 \cmidrule(lr){3-4}
 \cmidrule(lr){5-6}
 \cmidrule(lr){7-8}
 \cmidrule(lr){9-10}
 \cmidrule(lr){11-12}
{} & {} & $K=1$ & $K=5$ & $K=1$ & $K=5$ & $K=1$ & $K=5$ & $K=1$ & $K=5$ & \textbf{$K=1$} & \textbf{$K=5$}\\
   \midrule
   \multirow{5}{0.8cm}{\textbf{low}\\profile} 
       & Prototypical Network & 82.20 & 91.57 & 74.37 & 86.48 & 80.06 & 89.62 & 94.29 & 98.10 & 82.73 $\pm$ 2.32 & 91.44 $\pm$ 1.92 \\
        & ours w/ BT & 83.83 & 92.16 & \underline{78.70} & \underline{89.36} & 80.84 & 90.87 & 94.06 & 97.62 & 84.36
    $\pm$ 1.15
& 92
.50 $\pm$ 0.94 \\

        & ours w/ DBS & 83.10 & 92.56 & \underline{80.06} & \underline{90.21} & 82.31 & \underline{91.64} & 93.70 & 97.83 & 84.80 $\pm$ 1.26 & 93.06 $\pm$ 0.99 \\

        & ours w/ DBS+bigram  & 86.04 & 93.55 & \underline{82.09} & \underline{\textbf{91.57}} & \underline{83.60} &
\underline{92.71} & 95.11 & 98.23 & \underline{86.71 $\pm$ 1.14} & \underline{94.01 $\pm$ 1.05} \\

        & ours w/ DBS+unigram  & \underline{\textbf{87.23}} & \underline{\textbf{94.29}} & \underline{\textbf{83.70}}
& \underline{91.29} & \underline{\textbf{85.16}} & \underline{\textbf{93.00}} & \textbf{95.92} & \textbf{98.56} &
\textbf{\underline{88.00 $\pm$ 1.22}} & \textbf{\underline{94.29 $\pm$ 0.76}} \\

\addlinespace
\midrule
\addlinespace
  \multirow{5}{0.8cm}{\textbf{full} \\ profile}
  & Prototypical Network & 86.28 & 93.94 & 77.09 & 89.02 & 82.76 & 91.37 & 96.05 & 98.61 & 85.55 $\pm$ 2.20 & 93.24 $\pm$ 1.22 \\
   & ours w/ BT & 87.46 & 94.47 & 81.31 & 91.44 & 84.14 & 92.67 & 95.19 & 98.36 & 87.02 $\pm$ 1.36 & 94.23 $\pm$ 0.82 \\
   & ours w/ DBS & 86.94 & 94.50 & 82.35 & 91.68 & 84.42 & 92.62 & 94.85 & 98.41 & 87.14 $\pm$ 1.36 & 94.30 $\pm$ 0.60 \\
  & ours w/ DBS+bigram & 88.14 & 94.70 & 84.05 & 92.14 & 85.29 & 93.23 & 95.77 & 98.50 & 88.31 $\pm$ 1.43 & 94.64 $\pm$ 0.59 \\
  & ours w/ DBS+unigram & \textbf{89.56} & \textbf{94.71} & \textbf{84.34} & \textbf{92.55} & \textbf{86.11} & \textbf{93.70} & \textbf{96.49} & \textbf{98.74} & \textbf{89.13 $\pm$ 1.13} & \textbf{94.92 $\pm$ 0.57}  \\
 \bottomrule
\end{tabular}
}
\caption{5-way 1-shots and 5-way 5-shots accuracy on the test sets for each dataset. The \textit{ours} method is \protoaugment{} (unsupervised consistency loss using diverse paraphrases) equipped with different  paraphrasing strategies. For each dataset $\times$ C-way K-shot setting, we compute the average and the standard deviation over the $5$ runs (see Section~\ref{subsec:settings}), so that the last two columns contains average accuracy and $\pm$ the average standard deviations. For each data profile, we highlight the best method in \textbf{bold}. We \underline{underline} the methods on the \textbf{low} profile which perform better than the Prototypical Networks on the \textbf{full} profile. We trained $400$ different meta-learners -- $5$ methods, $2$ data profiles, $4$ datasets, $2$ meta-learning setup ($K={1,5}$) and $5$ runs for each configuration.
}
\label{tab:results}
\end{table*}
    \subsection{Intent detection results}\label{subsec:results-intent-detection}
    In this section, we discuss the accuracy results for the different meta-learners, for the standard 5-way and \{1, 5\}-shots meta-learning scenarios, as provided in Table~\ref{tab:results}. The reported metric is the accuracy on the test set at the iteration where the validation set's accuracy is maximal. Our DBS+unigram strategy row corresponds to the \texttt{flat} masking strategy, with $p_{\text{mask}}=0.7$. First, all methods augmented with unsupervised diverse paraphrasing outperform prototypical networks. However, back translation demonstrates only a limited improvement over the vanilla prototypical network due to their narrow diversity for short texts. Using paraphrases from DBS yields better results -- about 0.5 points over BT, on average --, hinting that using diverse paraphrases in the unsupervised consistency loss allows the few-shot model to build more robust sentence representations and therefore provides improved generalization capacities. Those results are consistent across the different datasets, except for Clinic for which accuracies are all very high, making all methods hardly separable. The dataset is not challenging enough, or in other words, meta-learning is robust to unbalanced short text classification problems given the nature of that dataset.

    These results illustrate the need for unsupervised paraphrasing and show that using diverse paraphrases
    provide a significant performance leap. In the 1-shot (resp. 5-shot) scenario, our best meta-learner improves
    prototypical networks by $5.27$ (resp. $2.85$) points on average.
    Remember that these improvements are made in an unsupervised manner hence at no additional cost.
    Slightly different from to~\cite{Xie20}, we do not find statistical differences depending on the rate at which
    $\tilde{L}$ is annealed in \protoaugment{} loss ($\alpha \in \{0.25, 1, 4\}$), which makes it easier to tune -- our unsupervised loss serves as a consistency regularization. Due to space limitations, this analysis is available in appendix~\ref{app:loss-annealing-strategy-appendix}.

    Adding our masking strategies on top of DBS has a significant impact on all datasets, with the unigram variant being up about 2 points over the vanilla DBS on average. On all datasets except Clinic, given only 10 labeled samples per class (\textbf{low} profile), it even outperforms the supervised baseline which is given the full training data (\textbf{full} profile). This means that \protoaugment{} does better than prototypical networks with much less -- $15$ times, and up to $47$ times, depending on the dataset -- labeled sentences per class. Those results indicate that our method more than compensates for the lack of labeled data and that no matter the amount of data available for the training class, there is a performance ceiling you cannot overcome without adding unsupervised knowledge from the validation and test classes. In the \textbf{full} profile, when given all the training data, our method greatly surpasses the Prototypical Network -- 3.58 points given 1 shot, on average. Moreover, \protoaugment{} is not only suited for the case where very little training data is available (\textbf{low} profile):  when sampling shots from the entire training dataset (\textbf{full} profile), it outperforms a fully supervised baseline. Furthermore, note that our method is consistently more stable than the supervised baselines, as its average standard deviation over the different runs is much lower than the vanilla Prototypical Network.

    \subsection{Masking strategies}\label{subsec:results-masking-strategies}

\begin{figure}[h]
        \centering
        \includegraphics[width=\linewidth]{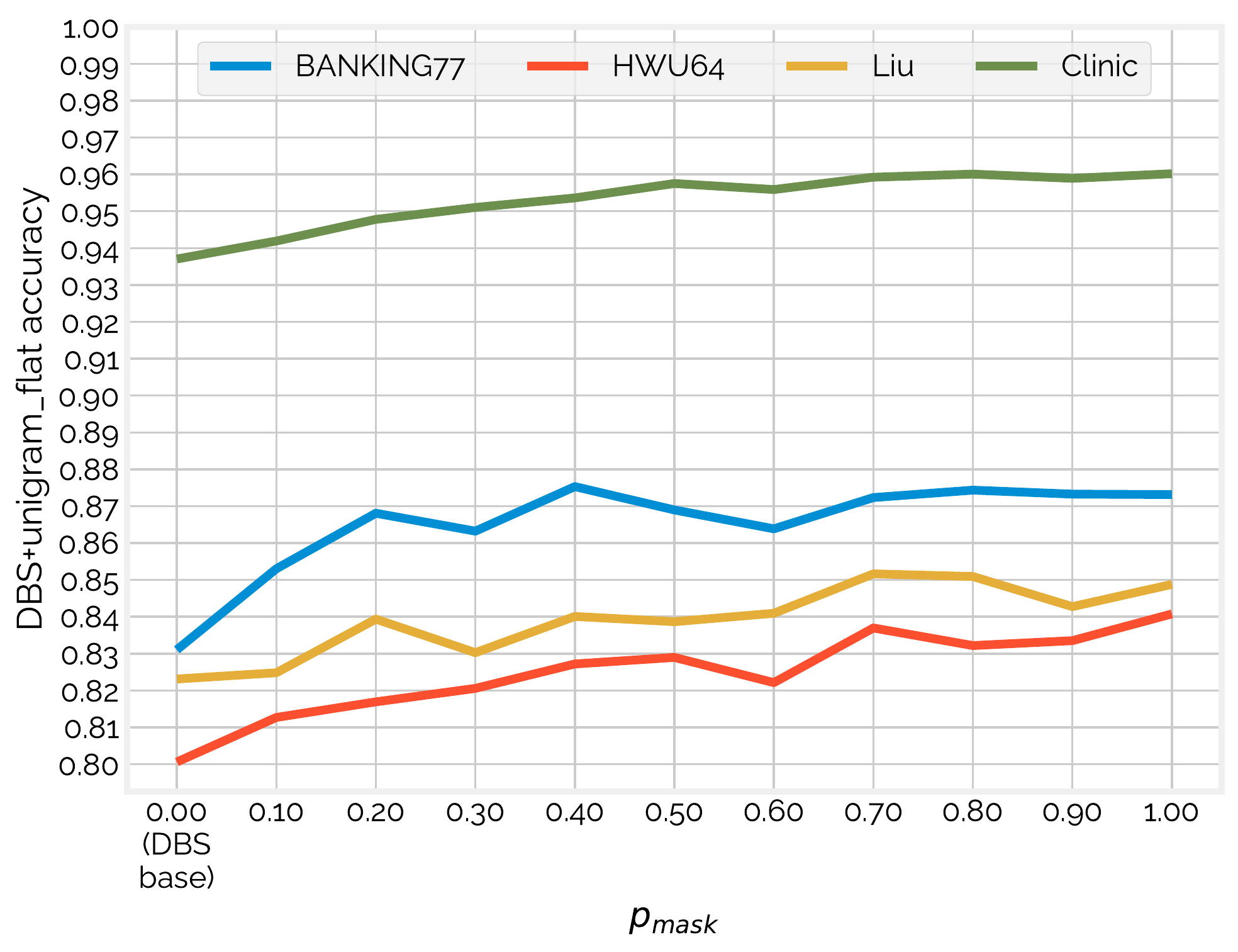}
        \caption{5-way 1-shot accuracy of DBS-unigram-flat method using different values of $p_{\text{mask}}$. Setting this value to 0 corresponds to the vanilla DBS without masking strategies.}
        \label{fig:results-pmax}
    \end{figure}

    We experimented with three variants of the unigram strategy (Section~\ref{subsec:pa:constraienddbs}), each assigning a different drop chance to each token depending on its position in the input sentence. In our experiments, we did not observe any significant difference in performance when putting more weight on the first tokens (\textit{down}), or last tokens (\textit{up}), or the same weight on all tokens (\textit{flat}) (Detailed results in appendix~\ref{app:masking-strategies-appendix}). We also conducted experiments where we tune the value $p_{\text{mask}}$, from $0$ to $1$, selecting $0.7$ as the best trade-off (Figure~\ref{fig:results-pmax}). This figure also clearly shows that the \textit{Clinic} dataset is one order of magnitude easier to solve than the other datasets.

    \section{Conclusion}\label{sec:conclusion}

%
In this work, we proposed \protoaugment{}, an architecture for meta-learning for the problem of classifying user-generated short-texts (intents).
%
%
%
%
We first introduced an unsupervised paraphrasing consistency loss in the prototypical network's framework to improve its representational power.
Then, while the recent diverse beam search algorithm was designed to enforce diversity between the generated paraphrases, it does not ensure diversity between the generated paraphrases and the original sentences. To make up for the latter, we introduce constraints in the diverse beam search generation, further increasing the diversity.
%
%
Our thorough evaluation demonstrates that \protoaugment{} offers a significant leap in accuracy for the most recent and challenging datasets. \protoaugment{} vastly outperforms prototypical networks, which was found to be the best meta-learning framework for short-texts~\cite{Dopierre21} against unsupervised-extended Prototypical Networks~\cite{protopp}, Matching Networks ~\cite{vinyals2016matching}, Relation Networks~\cite{Sung_2018_CVPR}, and Induction Networks~\cite{induction}, thereby making \protoaugment{} the new state-of-the-art for this task.
%
%
We provide the source code of \protoaugment{} as well as code for evaluations reported in this paper on a public repository~\footnote{https://github.com/tdopierre/ProtAugment}

\section*{Acknowledgments}
    We are thankful for the discussion we had with {Michele Bevilacqua, Marco Maru, and Roberto Navigli} from Sapienza university about diversity in Natural Language Generation. We also would like to thank ANRT~\footnote{https://www.anrt.asso.fr/fr} for making partnerships between companies and universities happen.


    \bibliographystyle{acl_natbib}
    \bibliography{acl2021}


\newpage
\appendix
\onecolumn
\FloatBarrier
\section{Diverse paraphrase samples}\label{app:diverse-paraphrase-samples}

 \begin{table*}[hbt!]
 \centering
\resizebox{0.7\linewidth}{!} {%
\begin{tabular}{l}
\toprule
\textit{orig}: Are you considered a bot?\\
\textit{back}: Do you consider yourself a bot?\\
\textit{dbs\_0}: Is there any way to be considered a bot?\\
\textit{dsb\_1}: Is there any way to know if someone is bot?\\
\textit{dbs\_2}: Is there any way to know if a person is bot?\\
\midrule


\textit{orig}: what is the square root of 31\\
\textit{back}: what is the square root of 31\\
\textit{dbs\_0}: How can I find square roots of 31 in mathematics?\\
\textit{dsb\_1}: How can I find quadratic roots of 31?\\
\textit{dbs\_2}: What is square roots and what is its effect on 31\\
\midrule


\textit{orig}: How long will my transfer be pending for?\\
\textit{back}: How long will my transfer be on hold?\\
\textit{dbs\_0}: How long will my transfer be pending? I am in first year.\\
\textit{dsb\_1}: When are all transfers coming up and how many days are they expected?\\
\textit{dbs\_2}: If I have a transfer for a while, how long should I wait for it?\\
\midrule

\textit{orig}: How to cancel a transaction\\
\textit{back}: How to cancel a transaction\\
\textit{dbs\_0}: What are some ways to cancel a transaction (in any way)?\\
\textit{dsb\_1}: What are some ways of cancelling an account (in some cases also involving transaction ).\\
\textit{dbs\_2}: Is there any way in which I can cancel an existing transaction?\\
\midrule

\textit{orig}: I am not sure where my phone is.\\
\textit{back}: I don't know where my phone is.\\
\textit{dbs\_0}: I am not really sure where my phone is located\\
\textit{dsb\_1}: How can I find the location of any Android mobile\\
\textit{dbs\_2}: I don't know where is my cell phone\\
\midrule

\textit{orig}: What do I need to do for a refund?\\
\textit{back}: What do I need to do to get a refund?\\
\textit{dbs\_0}: What should I do now that I am not getting a refund?\\
\textit{dsb\_1}: What are things that should be done before resumption of service ( reimbursed)?\\
\textit{dbs\_2}: What should you do if you want to refund your period of data?\\
\midrule

\textit{orig}: does pizza hut have my order ready\\
\textit{back}: Does Pizza Hut has my order ready\\
\textit{dbs\_0}: Does the pizza Hut have all your orders ready?\\
\textit{dsb\_1}: Does pizza Hut have all your orders made up?\\
\textit{dbs\_2}: Does the pizza Hut have all your orders ready for delivery?\\
\midrule

\textit{orig}: go silent for a day\\
\textit{back}: Quiet for a day.\\
\textit{dbs\_0}: Do you stay silent for a day or go silent for another day\\
\textit{dsb\_1}: To the rest of the day, stay peaceful and collected.\\
\textit{dbs\_2}: So, to the rest of the day, go silent only.\\
\midrule

\textit{orig}: what's the recipe for fish soup\\
\textit{back}: What is the recipe for fish soup\\
\textit{dbs\_0}: How do you make fish soup? How is the recipe determined?\\
\textit{dsb\_1}: How can you recipe for fish-sugary food?\\
\textit{dbs\_2}: What are the recipes for Fish soup and how is it prepared?\\
\midrule

\textit{orig}: Find easy recipe for almond milk\\
\textit{back}: Find an easy recipe for almond milk\\
\textit{dbs\_0}: What are some good recipe for Almond milk?\\
\textit{dsb\_1}: What are some good ways of making Almond milk?\\
\textit{dbs\_2}: How do I make Almond milk for a beginner?\\
\midrule

\textit{orig}: Will I need to wear a coat today?\\
\textit{back}: Should I wear a coat today?\\
\textit{dbs\_0}: Today, do I need to put on a coat\\
\textit{dsb\_1}: Should I wear a coat and what kind of coat\\
\textit{dbs\_2}: What should I wear to work today, and why\\
\midrule

\textit{orig}: can you play m3 file\\
\textit{back}: can you read m3 file\\
\textit{dbs\_0}: M3 files: can I play the entire M3 file?\\
\textit{dsb\_1}: Is there any way to play 3M files on Earth without downloading it\\
\textit{dbs\_2}: Is there any way to play M3 files on Windows?\\

\bottomrule
\end{tabular}
}
\caption{Additional paraphrases samples. }
\label{tab:samples-appendix}
\end{table*}




\section{Paraphrase Diversity Evaluation}\label{app:paraphrase-diversity-evaluation}

\begin{table*}[hbt!]
    \centering
    \resizebox{\linewidth}{!}{
    \begin{tabular}{@{\extracolsep{1pt}}l
    ccc
    ccc
    ccc
    ccc}
    \toprule
         &  \multicolumn{3}{c}{BANKING77} & \multicolumn{3}{c}{HWU64} & \multicolumn{3}{c}{Liu} & \multicolumn{3}{c}{Clinic}\\
         \cmidrule(lr){2-4}
         \cmidrule(lr){5-7}
         \cmidrule(lr){8-10}
         \cmidrule(lr){11-13}
         & \textbf{BLEU} & \textbf{dist-2}  & \textbf{use}
         & \textbf{BLEU} & \textbf{dist-2}  & \textbf{use}
         & \textbf{BLEU} & \textbf{dist-2}  & \textbf{use}
         & \textbf{BLEU} & \textbf{dist-2}  & \textbf{use}\\
         \midrule
         back-translation & 56.0 & 0.183 & 0.896 & 40.2 & 0.307 & 0.888 & 47.7 & 0.268 & 0.892 & 43.9 & 0.205 & 0.903 \\
         DBS & 34.2 & 0.200 & 0.807 & 19.5 & 0.340 & 0.769 & 19.7 & 0.293 & 0.750 & 22.3 & 0.236 & 0.805 \\
         DBS+bigram & 0.1 & 0.228 & 0.702 & 0.1 & 0.350 & 0.692 & 0.4 & 0.293 & 0.664 & 0.2 & 0.257 & 0.717 \\
         DBS+unigram & 0.2 & 0.343 & 0.613 & 0.5 & 0.407 & 0.628 & 0.5 & 0.351 & 0.596 & 0.3 & 0.323 & 0.644 \\
    \bottomrule
    \end{tabular}}
    \caption{Paraphrase evaluation on all 4 datasets. The unigram variant exposed here is using the \textit{flat} masking strategy with $p_{\text{mask}}=0.7$.}
    \label{tab:paraphrase-evaluation-appendix}
\end{table*}

\section{Masking tokens depending on their position}\label{app:masking-strategies-appendix}

\begin{table*}[hbt!]
\resizebox{\linewidth}{!} {%
\begin{tabular}{@{\extracolsep{1pt}}lcccccccccc}
\toprule
{} & \multicolumn{8}{c}{\textbf{Datasets}} & \multicolumn{2}{c}{\textbf{Accuracy stats}}\\
 \cmidrule(lr){2-11}

  {\multirow{2}{1.1cm}{\textbf{Method}}} & \multicolumn{2}{c}{Banking} & \multicolumn{2}{c}{HWU} & \multicolumn{2}{c}{Liu} & \multicolumn{2}{c}{Clinic} & \multicolumn{2}{c}{(\textbf{$AVG\pm STD$})}\\
 \cmidrule(lr){2-3}
 \cmidrule(lr){4-5}
 \cmidrule(lr){6-7}
 \cmidrule(lr){8-9}
 \cmidrule(lr){10-11}
 {} & $K=1$ & $K=5$ & $K=1$ & $K=5$ & $K=1$ & $K=5$ & $K=1$ & $K=5$ & \textbf{$K=1$} & \textbf{$K=5$}\\
 \midrule
 DBS+unigram-\textit{flat} & 87.23 & 94.29 & 83.70 & 91.29 & 85.16 & 93.00 & 95.92 & 98.56 & 88.00 $\pm$ 1.22 & 94.29 $\pm$ 0.76 \\
DBS+unigram-\textit{down} & 87.43 & 94.14 & 83.06 & 92.14 & 84.87 & 93.33 & 95.93 & 98.61 & 87.82 $\pm$ 0.84 & 94.55 $\pm$ 0.71 \\
DBS+unigram-\textit{up} & 86.18 & 94.12 & 83.30 & 91.21 & 85.14 & 93.15 & 95.84 & 98.30 & 87.62 $\pm$ 1.23 & 94.20 $\pm$ 0.70 \\
\bottomrule

\end{tabular}
}
\caption{Performances of DBS+unigram strategies putting either more chance to mask first tokens (\textit{down}), last tokens (\textit{up}), or the same chance to all tokens (\textit{flat}). All strategies use $p_{\text{mask}}=0.7$. Overall, there is no significant difference between the three strategies.}
\label{tab:results-masking-strategies}
\end{table*}

\section{Loss annealing strategy}\label{app:loss-annealing-strategy-appendix}

\begin{table*}[hbt!]
\resizebox{\linewidth}{!} {%
\begin{tabular}{@{\extracolsep{1pt}}lccccccccccc}
\toprule
{} & {} & \multicolumn{8}{c}{\textbf{Datasets}} & \multicolumn{2}{c}{\textbf{Accuracy stats}}\\
 \cmidrule(lr){3-12}

  {\multirow{2}{1.1cm}{\textbf{Method}}} & & \multicolumn{2}{c}{Banking} & \multicolumn{2}{c}{HWU} & \multicolumn{2}{c}{Liu} & \multicolumn{2}{c}{Clinic} & \multicolumn{2}{c}{(\textbf{$AVG\pm STD$})}\\
 \cmidrule(lr){3-4}
 \cmidrule(lr){5-6}
 \cmidrule(lr){7-8}
 \cmidrule(lr){9-10}
 \cmidrule(lr){11-12}
 {} & $\alpha$ & $K=1$ & $K=5$ & $K=1$ & $K=5$ & $K=1$ & $K=5$ & $K=1$ & $K=5$ & \textbf{$K=1$} & \textbf{$K=5$}\\
 \midrule
 \multirow{3}{3cm}{DBS+unigram-\textit{flat}} & 1 & 87.23 & 94.29 & 83.70 & 91.29 & 85.16 & 93.00 & 95.92 & 98.56 & 88.00 $\pm$ 1.22 & 94.29 $\pm$ 0.76 \\
 & 0.25 & 86.71 & 94.17 & 82.71 & 91.19 & 85.52 & 93.11 & 95.99 & 98.44 & 87.73 $\pm$ 1.09 & 94.23 $\pm$ 0.85 \\
 & 4 & 86.90 & 94.14 & 83.26 & 92.35 & 84.48 & 93.17 & 95.69 & 98.49 & 87.58 $\pm$ 1.64 & 94.54 $\pm$ 0.81 \\
\bottomrule

\end{tabular}
}
\caption{Performances of DBS+unigram strategies with different values of the loss annealing parameter $\alpha$. All strategies use $p_{\text{mask}}=0.7$. Overall, there is no significant difference when changing the value of $\alpha$.}
\label{tab:results-masking-strategies}
\end{table*}

\end{document}